\documentclass[10pt,twocolumn,letterpaper]{article}

\usepackage{wacv}
\usepackage{times}
\usepackage{epsfig}
\usepackage{graphicx}
\usepackage{amsmath}
\usepackage{amssymb}
\usepackage{bbm}
\usepackage{bbold}
\def\wacvPaperID{****} 
\usepackage{lipsum}
\usepackage{mwe}

\ifwacvfinal
\else
\fi

\def\httilde{\mbox{\tt\raisebox{-.5ex}{\symbol{126}}}}

\usepackage{amsmath}
\usepackage{amssymb}
\usepackage{wacv}
\usepackage[utf8]{inputenc} 
\usepackage[T1]{fontenc}    
\usepackage{url}            
\usepackage{booktabs}       
\usepackage{amsfonts}       
\usepackage{nicefrac}       
\usepackage{microtype}      

\usepackage{amsmath}
\usepackage{bbm}
\usepackage{epsfig}
\usepackage{algorithm}
\usepackage{algorithmic}
\usepackage{subfigure}
\usepackage{epstopdf}
\usepackage{booktabs}
\usepackage{array}
\usepackage{multirow}
\usepackage{hhline}
\usepackage{makecell}
\usepackage{framed}
\usepackage{enumitem}
\usepackage{wrapfig} 
\usepackage{float}
\usepackage{eso-pic}
\usepackage{xspace}
\usepackage{xcolor}
\usepackage[export]{adjustbox}
\usepackage{stackengine}
\newcommand{\R}{\mathbb{R}}

\usepackage{lscape}
\usepackage{wacv}
\usepackage{times}
\usepackage{epsfig}
\usepackage{graphicx}
\usepackage{amsmath}
\usepackage{amssymb}
\usepackage{amsfonts}
\usepackage{caption}

\def\wacvPaperID{55} 

\wacvfinalcopy 
\author{Abhinav Narayan Harish\\
Indian Institute of Technology Gandhinagar\\
{\tt\small abhinav.narayan@iitgn.ac.in}

\and
Rajendra Nagar\\
Indian Institute of Technology Jodhpur\\
{\tt\small rn@iitj.ac.in}
\and
Shanmuganathan Raman\\
Indian Institute of Technology Gandhinagar\\
{\tt\small shanmuga@iitgn.ac.in}
}

\ifwacvfinal
\usepackage[breaklinks=true,bookmarks=false]{hyperref}
\else
\usepackage[pagebackref=true,breaklinks=true,colorlinks,bookmarks=false]{hyperref}
\fi

\title{RGL-NET: A Recurrent Graph Learning framework for Progressive Part Assembly
}

\def\httilde{\mbox{\tt\raisebox{-.5ex}{\symbol{126}}}}
\begin{document}


\maketitle
    

\begin{abstract}
Autonomous assembly of objects is an essential task in robotics and 3D computer vision. It has been studied extensively in robotics as a problem of motion planning, actuator control and obstacle avoidance. However, the task of developing a generalized framework for assembly robust to structural variants remains relatively unexplored. In this work, we tackle this problem using a recurrent graph learning framework considering inter-part relations and the progressive update of the part pose.  Our network can learn more plausible predictions of shape structure by accounting for priorly assembled parts.
Compared to the current state-of-the-art, our network yields up to 10\% improvement in part accuracy and up to 15\% improvement in connectivity accuracy on the PartNet~\cite{mo2019partnet} dataset. Moreover, our resulting latent space facilitates exciting applications such as shape recovery from the point-cloud components. We conduct extensive experiments to justify our design choices and demonstrate the effectiveness of the proposed framework.


\end{abstract}

\section{Introduction}
\label{sec:intro}
\par
Automated assembly requires a structural and functional understanding of object parts to place them in their appropriate locations. In a chair, a square-shaped structure could be its base or its back. A long cuboid part could be its legs. However, imparting this assembly skill to machines is still an open problem in vision and robotics. 
\par
\begin{figure}[!h]
    \begin{center}
    \stackunder{\includegraphics[width=0.26\linewidth]{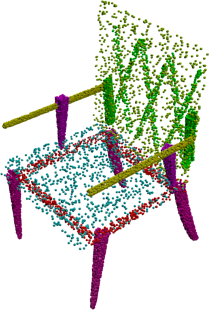}}{B-DGL}\hspace{7mm}
    \stackunder{\includegraphics[width=0.26\linewidth]{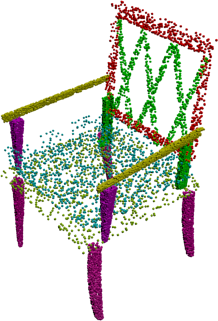}}{Ours}\hspace{7mm}
    \stackunder{\includegraphics[width=0.26\linewidth]{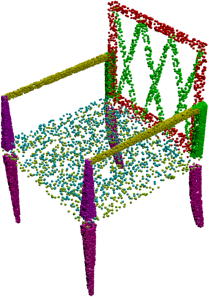}}{Ground-truth}\hspace{7mm}
    \end{center}    
    \caption{Our progressive Part Assembly scheme reduces inter-part confusion. Dynamic Graph Learning~\cite{HuangZhan2020PartAssembly} (B-DGL) mistakes the green coloured chair seat for its back.}
    \label{fig:inter-part-confusion}
\end{figure}

To ensure smooth and collision-free assembly, we must  accurately estimate the pose of each part. In robotics literature, there are a few works that attempt this problem. Choi \etal~\cite{choi2012voting} develop a pose estimation scheme to register point cloud to incomplete depth maps. Suarez \etal~\cite{suarez2018can} assemble an IKEA chair by hard-coding motion trajectories onto robotic arm manipulators. However, none of the prior works can be generalized to household assembly tasks where we may not have access to the global structure of the assembled shape. In this work, we assemble a shape from its part point clouds without any prior semantic knowledge. Instead of manually configuring per-part pose, we explore relations that can be generalized across shapes in a category.
\par
A study conducted in 2003 on designing assembly instructions~\cite{agrawala2003designing} uncovers that humans prefer sequential assembly instructions - split into step-by-step instructions. This could be an assembly diagram illustrating how each part connects with the other or an instructional video. However, designing
detailed diagrams can become cumbersome for the designer. In some instances, intricate designs are often unnecessary. Understanding the assembly progression can provide information of the subsequent part poses. This work demonstrates that a linear ordering of part elements can significantly improve part placement and inter-part connectivity.
\par
Being a recently formulated research problem, only a few works tackle this problem in a similar setting as ours. Li \etal~\cite{li2020learning} assemble a shape from its component point cloud using an image-based prior. In Coalesce~\cite{yin2020coalesce}, the authors develop a framework for assembly and joint synthesis using translation and scaling of component point clouds. Huang \etal~\cite{HuangZhan2020PartAssembly} attempt this task without semantic knowledge of parts using a dynamic graph learning framework.
\par
\begin{figure}[t!]
    \centering
    \includegraphics[width = 1\linewidth]{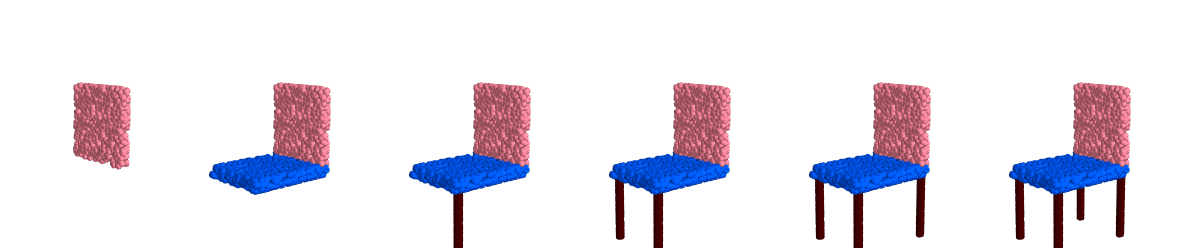}
    \caption{Top-down order for a chair in PartNet~\cite{mo2019partnet}.}
    \label{fig:canonical}
\end{figure}

However, none of these prior works have explored  progressive assembly strategies. They transform all parts at once without leveraging information that previous part placements can offer. This can result in confusion among structurally similar components. For instance, a chair seat may have a very similar structure to the chair back, resulting in its incorrect placement (Figure~\ref{fig:inter-part-confusion}). By incorporating an assembly progression (Figure~\ref{fig:canonical}), we can reduce inter-part confusion and increase the network's capacity to learn intricate structures. We encode this information in the hidden state of a recurrent neural network. 
\par
 Similar to~\cite{HuangZhan2020PartAssembly}, we account for structural variety by incorporating random noise and allowing our network to explore the ground truth space using the minimum-over-N (MoN)~\cite{fan2017point} loss. Further, we analyze our network performance at various dimensions of random noise. Our analysis reveals that our framework can generalize well even at the zero randomness setting. Overall, our progressive scheme demonstrates up to 10\% improvement in part accuracy and up to 15\% improvement in connectivity accuracy over dynamic graph learning~\cite{HuangZhan2020PartAssembly} on PartNet~\cite{mo2019partnet}. Moreover, our standalone framework can achieve up to 6\% improvement over this baseline, demonstrating its efficacy. Our ablation studies address the critical aspects of our scheme, including the architectural design and the optimal order for part placement.

In summary, our major contributions are - 
\begin{itemize}
    \itemsep0em 
    \item   We propose a novel recurrent graph learning framework for assembly which significantly improves part-placement and inter-part connectivity.
    \item Our framework yields competitive performance  even in the absence of random exploration.
    \item We demonstrate qualitatively the potency of our latent space by utilizing it to recover shape without access to its global structure. 
    \item We investigate a variety of ways of ordering part components, and experimentally establish the optimality of our choice.
\end{itemize}
\begin{figure*}[htbp]
    \centering
    \includegraphics[width=1\linewidth]{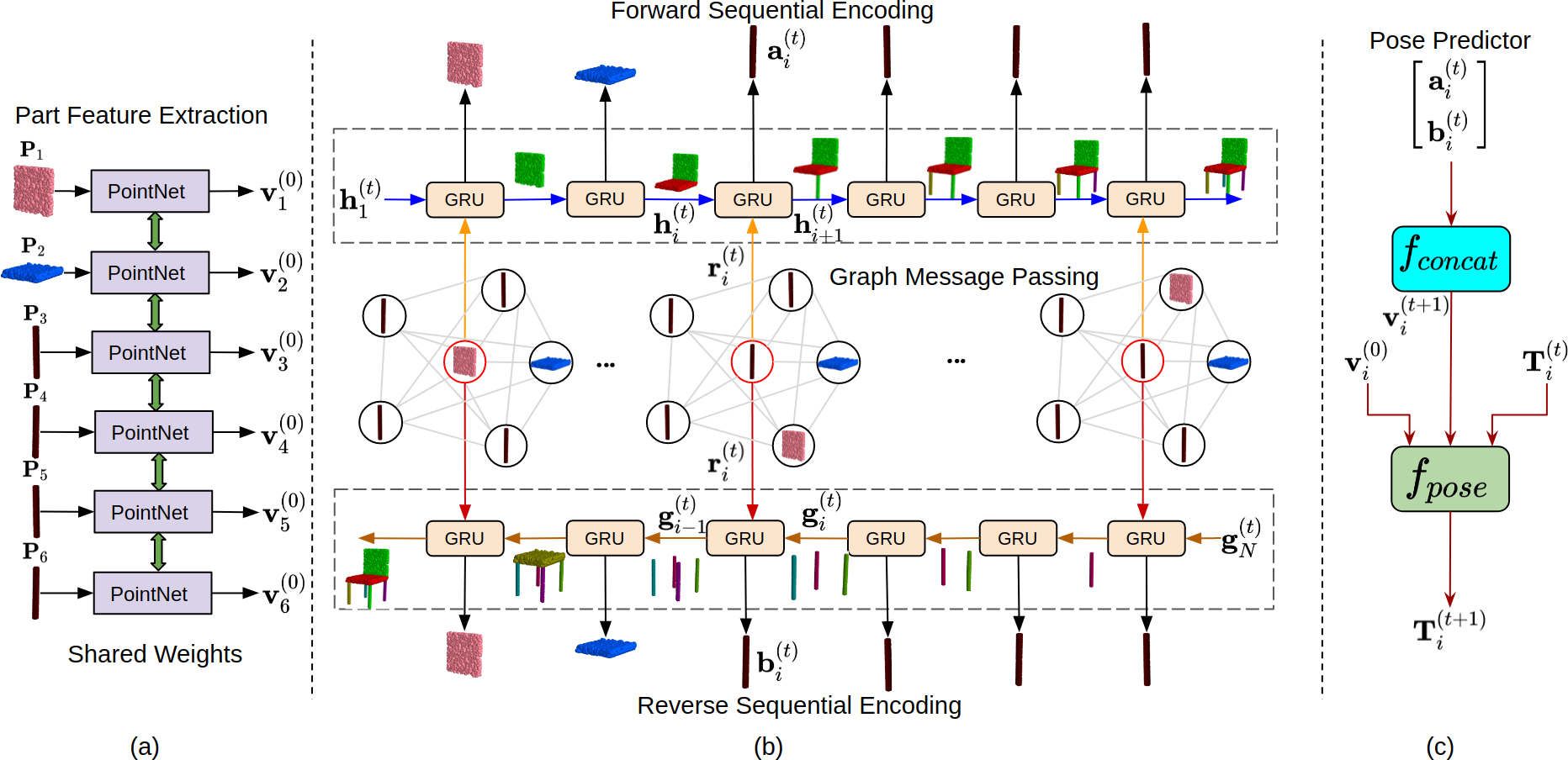}
    \caption{One iteration of our Recurrent Graph Learning framework. (a) We process part features and compute a graph message. (b) The message is encoded sequentially in our bidirectional GRU framework. (c) The features generated by the forward and reverse GRU are used to regress part-pose. We use three such iterations in our framework.}
    \label{fig:pipeline} 
\end{figure*}

\section{Related Work}
\textbf{Part Based 3D Modelling.}
We can decompose complex 3D shapes into simple part structures, which can construct novel shapes. One of pioneering works in this direction was by Funkhouser \etal~\cite{funkhouser2004modeling}, who attempted this problem using an intelligent scissoring of parts components. The subsequent works~\cite{chaudhuri2011probabilistic, kalogerakis2012probabilistic, jaiswal2016assembly} utilize probabilistic graphical models to encode semantic part relationships. The authors of~\cite{chaudhuri2010data} demonstrate the construction of high-quality CAD models using noisy data from sensors and a 3D shape database.
\par
Recent works leverage the power of deep neural networks for shape-modelling. ComplementMe~\cite{sung2017complementme} introduces a weakly supervised approach in the absence of consistent semantic segmentation and labels. The authors of ~\cite{dubrovina2019composite} create an autoencoder for a latent space to factorize a shape into its parts, allowing for part-level shape manipulation.
\par
Most of the prior works in this domain either assume known part semantics or depend on an existing shape repository. We make no such assumption and assemble a variable number of parts during testing. 

\textbf{Structural Shape Generation.} 
With the advent of deep-learning and the development of large scale shape datasets~\cite{mo2019partnet, yi2016scalable}, shape generation has garnered the interest of the vision community. GRASS~\cite{li2017grass} and  StructureNet~\cite{mo2019structurenet} compress shape structure into a latent space, taking into account inter-part relationships. PT2PC~\cite{mo2020pt2pc} generates 3D shapes conditioned on the part-tree decomposition. ShapeAssembly~\cite{jones2020shapeassembly} uses a procedural programmatic representation for connecting part cuboids. SAGNET~\cite{wu2019sagnet} develops a structural aware generative model, catering to pairwise relationships and encoding structure and geometry separately. SDM-NET~\cite{gao2019sdm} extends this approach to meshes through a controlled generation of fine-grained geometry.
\par
Few of these prior works model shape generation as an assembly of point cloud components. Inspired by Seq2Seq networks for machine translation, PQ-NET~\cite{wu2020pq} develops a sequential encoding and decoding scheme for regressing shape parameters. PageNet~\cite{li2020pagenet} utilizes a partwise-VAE to regress the transformation parameters of a 3D shape.
\par
Instead of generating a new point cloud structure, we transform the existing point clouds of shape components using a rigid transformation. Our problem setting is more challenging, as we lack access to the assembled shape, and is more relevant to real-world vision and robotics applications.  
\par
\textbf{Part Component Assembly.} 
Automated part assembly is a long-standing problem in robotics, emphasizing 6D pose estimation, motion planning and actuator control. Shao \etal ~\cite{shao2020learning} utilize fixtures to reduce the complexity of the assembly space. Zakka \etal~\cite{zakka2020form2fit}  generalize assembly to unseen categories using shape descriptors. The authors of~\cite{luo2019reinforcement} utilize reinforcement learning to incorporate parameters like force and torque into assembly. Several other works formulate assembly as a motion planning problem~\cite{hutchinson1990extending, jimenez2013survey}. 
\par
We tackle the problem closely aligned to computer vision, wherein we estimate the 6D pose from part point clouds without prior semantic knowledge. In this domain, ~\cite{li2020learning, HuangZhan2020PartAssembly} formulate a similar problem to ours. Li \etal~\cite{li2020learning} utilize a two-stage pipeline of image segmentation followed by part assembly. The authors of~\cite{HuangZhan2020PartAssembly} utilize a dynamic graph framework to assemble a shape. However, unlike these prior works, we incorporate progressive assembly to encode information, significantly improving part-placement. 
\section{Proposed Method}
Consider an ordered set of $N$ point clouds components of a 3D shape, $\mathcal{P} = (\mathbf{P}_{1}, \mathbf{P}_{2}, \ldots, \mathbf{P}_{N})$, where $\mathbf{P}_{i} \in \mathbb{R}^{N_{d} \times 3}$, and $N_{d}$, represents the number of points per 3D shape. We predict part poses $(\mathbf{q}_{i}, \mathbf{c}_{i})$, where, $\mathbf{q}_{i} \in \mathbb{R}^{4}$ given $\|\mathbf{q}_i\|_{2} = 1$ represents the  quaternion and $\mathbf{c}_{i} \in \mathbb{R}^{3}$ represents the translation. 
 The complete assembled shape is $\mathcal{S} = \mathbf{T}_{1}(\mathbf{P}_{1})\cup \mathbf{T}_{2}(\mathbf{P}_{2})\cup\cdots\cup \mathbf{T}_{N}(\mathbf{P}_{N})$. Here, $\mathbf{T}_{i}(.)$ represents joint $SE(3)$ transformation arising from  $(\mathbf{q}_{i}, \mathbf{c}_{i})$. 
 \par
To assemble a shape, we utilize an iterative network composed of a graph neural network backbone~\cite{HuangZhan2020PartAssembly} and a progressive assembly encoder. The graph neural network backbone accounts for inter-part relations to comprehend contextual information. Progressive encoding accumulates a prior using the shape structure of already assembled parts. We provide the complete pipeline of our framework in Figure~\ref{fig:pipeline}.   
\subsection{Graph Learning Backbone}\label{GLB}

We model the inter-part relations, using a time-varying dynamic graph with set of vertices $\mathcal{V}^{(t)}$ and edges $ \mathcal{E}^{(t)}$. The nodes of the graph $\mathcal{V}^{(t)}=\{\mathbf{v}_{1}^{(t)},\mathbf{v}_{2}^{(t)},\ldots,\mathbf{v}_{N}^{(t)}\}$ are the features of each part $\mathbf{P}_{i}$ at time step $t$ of the iterative network. The graph is complete with a self-loop, i.e., ${(i,j) \in \mathcal{E}^{(t)}} \hspace{1mm} \forall \; (i,j)\in[N]\times[N]$. Here, $[N]$ denotes the set of first $N$ natural numbers $\{1, 2,  \ldots, N\}$. We initialize the features $\mathbf{v}_i^{(0)}\in \R^{256}$ using a shared PointNet~\cite{qi2017pointnet} encoder on the point-cloud $\mathbf{P}_{i}$.  At time step $t$, we model the edge message $\mathbf{e}_{ij}^{(t)}\in\mathbb{R}^{256}$ between the $i$-th and $j$-th nodes as, 
\begin{equation}
\mathbf{e}_{i j}^{(t)}=f_{\text {edge }}\left(\begin{bmatrix}\mathbf{v}_{i}^{(t)}\\ \mathbf{v}_{j}^{(t)}\end{bmatrix}\right).
\end{equation}

During assembly, distinct part-pairs may bear a different relationship. For instance, the four legs of a chair could be strongly dependent on each other and less influenced by the position of the chair back. To account for this, we use an attention mechanism~\cite{vaswani2017attention}. Accordingly, we compute the overall message received by $\mathbf{v}_{i}^{(t)}$  as a weighted combination of edge messages from all possible nodes $\mathbf{v}_{j}^{(t)}$.
\par
\begin{equation}
    \mathbf{m}_{i}^{(t)} = \frac{\sum_{j=1}^{N}w_{ij}^{(t)}\mathbf{e}_{i j}^{(t)}}{\sum_{j=1}^{N}w_{ij}^{(t)}}.
\end{equation}
Here, $w_{ij}^{(t)}$ represents the scalar attention weight between nodes $\mathbf{v}_{i}^{(t)}$ and $\mathbf{v}_{j}^{(t)}$. Among the many possible ways to compute attention, we observe that using features extracted from part poses $\mathbf{T}_{i}^{(t)}$ and $\mathbf{T}_{j}^{(t)}$ yield good results.
\begin{equation}
    w_{ij}^{(t)} = f_{rel}(f_{feat}(\mathbf{T}_{i}^{(t)}), f_{feat}(\mathbf{T}_{j}^{(t)})), \hspace{1mm} \forall  t > 0.
\end{equation}
Here, $f_{feat}$ processes part-poses and returns a 128D feature. $f_{rel}$ operates on these features to return the scalar $w_{ij}$. At the initial time step, $w_{ij}^{(0)} = 1$ and  $\mathbf{T}_{k}^{(0)}\left(\mathbf{P}_{k}\right) = \mathbf{P}_{k}, \hspace{1mm} \forall   k \in [N]$.
\subsection{Progressive Message Encoding}~\label{SeqEnc}
We identified two choices for progressive assembly - a) update the part features one at a time and use the updated features for relational reasoning with subsequent parts, b) storing the assembly information in a recurrent unit. We reject the first option because - i) we face the problem of vanishing and exploding gradients for parts occurring at the beginning of the sequence, ii) the parts at the end receive more supervision than the parts occurring at the beginning. Instead, we utilize a bidirectional gated recurrent unit (GRU) to store the prior assembly. This ensures smoother gradient flow. Moreover, its bidirectional nature distributes information fairly across the sequence. 

We model the network by two ordered sets of hidden states $\mathcal{H}^{(t)}=\{\mathbf{h}_{1}^{(t)},\mathbf{h}_{2}^{(t)},\ldots,\mathbf{h}_{N}^{(t)}\}$ and $\mathcal{G}^{(t)}=\{\mathbf{g}_{1}^{(t)},\mathbf{g}_{2}^{(t)},\ldots,\mathbf{g}_{N}^{(t)}\}$ for the forward and backward recurrent units, respectively. Here, $\mathbf{h}_k^{(t)},\mathbf{g}_k^{(t)}\in \mathbb{R}^{256},\; \forall k\in[N]$. We allow our network to explore the  ground truth space by encoding noise in the initial hidden state. 
\begin{align}
        \mathbf{h}_{1}^{(t)} = \mathbf{g}_{N}^{(t)} = \begin{bmatrix}\mathbf{z}^\top& \mathbf{0}^\top\end{bmatrix}^\top.
        \label{eq:initial_hidden}
\end{align}
\par
Here, $\mathbf{z} \sim \mathcal{N}(\mathbf{0}, \mathbf{I})$ represents the random noise vector. We keep the initial forward and reverse hidden states the same so that both learn similar shape structures. While regressing the part pose of a current part $\mathbf{P}_{i}$, we take into account its current features and the received part message. The recurrent input,  $\mathbf{r}_{i}^{(t)} =\begin{bmatrix}\mathbf{v}_{i}^{(t)}\\ \mathbf{m}_{i}^{(t)}\end{bmatrix}$ gives the network a context of the relative and absolute orientation of each part. We incorporate this information onto the prior using $f_{hidden}$.
\begin{align}
    \mathbf{h}_{i+1}^{(t)} &= f_{hidden}(\mathbf{r}_{i}^{(t)}, \mathbf{h}_{i}^{(t)})\\
    \mathbf{g}_{i-1}^{(t)} &= f_{hidden}(\mathbf{r}_{i}^{(t)}, \mathbf{g}_{i}^{(t)}).
    \label{eq:update}
\end{align}
Correspondingly, for each part we obtain two outputs, $\mathbf{a}_{i}^{(t)}$ and $\mathbf{b}_{i}^{(t)}$ through forward and reverse encoding, respectively. 
\begin{align}
    \mathbf{a}_{i}^{(t)} &= f_{out}(\mathbf{r}_{i}^{(t)}, \mathbf{h}_{i}^{(t)}) \\
    \mathbf{b}_{i}^{(t)} &= f_{out}(\mathbf{r}_{i}^{(t)}, \mathbf{g}_{i}^{(t)}).
\end{align}
\par
We model the updated features $\mathbf{v}_{i}^{(t + 1)}$ by processing  $\mathbf{a}_{i}^{(t)}$, $\mathbf{b}_{i}^{(t)} \in \mathbb{R}^{256}$ using a function $f_{concat}$.
\begin{equation}
    \mathbf{v}_{i}^{(t+1)} = f_{concat}\left(\begin{bmatrix}\mathbf{a}_{i}^{(t)}\\ \mathbf{b}_{i}^{(t)}\end{bmatrix}\right).
\end{equation}
\par
This step aims to reduce the bias occurring due to part location in the sequence; parts appearing at the beginning of the first sequence would occur at the end of the second and vice-versa. Using these updated features, we can regress the pose for each part. We also utilize the original features $\mathbf{v}_{i}^{(0)}$ and previously extracted part-pose
$\mathbf{T}_{i}^{(t)}$, to pass on information extracted in previous time-steps.  
\begin{equation}
    \mathbf{T}_{i}^{(t+1)} = f_{pose}(\mathbf{v}_{i}^{(t + 1)}, \mathbf{v}_{i}^{(0)}, \mathbf{T}_{i}^{(t)}).
\end{equation}
In our implementation, $f_{out}$ and $f_{hidden}$ are the transfer functions of the GRU block. $f_{rel}$, $f_{edge}$, $f_{feat}$, $f_{pose}$ and $f_{concat}$ are parameterized by Multi-Layer-Perceptrons (MLP's).  Overall, we utilize three time steps of graph encoding and progressive assembly. 

\definecolor{darkgreen}{RGB}{0,120,0}
\setlength{\tabcolsep}{4pt}
\begin{table*}[!h]
		\small
		\begin{center}
			\begin{tabular}{c|c|c|c|c|c|c|c}
             \hline
             \multicolumn{1}{c}{} &  & B-Global~\cite{schor2019componet, li2020pagenet} & B-LSTM~\cite{wu2020pq} & B-Complement~\cite{sung2017complementme} & B-DGL~\cite{HuangZhan2020PartAssembly} & Ours without MoN & Ours (Complete) \\ \hline
             \multirow{3}{*}{SCD$\downarrow$} & Chair & 0.0146 & 0.0131 & 0.0241 & \textbf{\textcolor{blue}{0.0091}} & 0.0101 & \textbf{\textcolor{darkgreen}{0.0087}} 
             \\
             \cline{3-8}
             & Table & 0.0112 & 0.0125 & 0.0298 & \textbf{\textcolor{blue}{0.0050}} & 0.0053 & \textbf{\textcolor{darkgreen}{0.0048}}
             \\
             \cline{3-8}
             & Lamp & 0.0079 & \textbf{\textcolor{blue}{0.0077}} & 0.0150 & 0.0093 & 0.0088 & \textbf{\textcolor{darkgreen}{0.0072}} \\
             \hline
             \multirow{3}{*}{PA$\uparrow$} & Chair & 15.70 & 21.77 & 8.78 & 39.00 & \textbf{\textcolor{blue}{42.84}} & \textbf{\textcolor{darkgreen}{49.06}} \\
             \cline{3-8}
             & Table & 15.37 & 28.64 & 2.32 & \textbf{\textcolor{blue}{49.51}} & 49.15 & \textbf{\textcolor{darkgreen}{54.16}}\\
             \cline{3-8}
             & Lamp & 22.61 & 20.78 & 12.67 & \textbf{\textcolor{blue}{33.33}} & 31.66 & \textbf{\textcolor{darkgreen}{37.56}} \\
             \hline
             \multirow{3}{*}{CA$\uparrow$} & Chair & 9.90 & 6.80 & 9.19 & 23.87 & \textbf{\textcolor{blue}{28.74}} & \textbf{\textcolor{darkgreen}{32.26}} \\
             \cline{3-8}
             & Table & 33.84 & 22.56 & 15.57 & \textbf{\textcolor{blue}{39.96}} & 39.71 & \textbf{\textcolor{darkgreen}{42.15}}\\
             \cline{3-8}
             & Lamp & 18.60 & 14.05 & 26.56 & 41.70 & \textbf{\textcolor{blue}{46.28}} &\textbf{\textcolor{darkgreen}{57.34}} \\
             \hline
            \end{tabular}
		\end{center}
		\caption{Quantitative comparison with baseline methods. Here SCD: Shape Chamfer Distance, PA: Part Accuracy and CA: Connectivity Accuracy. \textbf{\textcolor{darkgreen}{Green}} represents the best performance and \textbf{\textcolor{blue}{Blue}} represents the second best. }
		\label{table:performance}
\end{table*}

\section{Experiments}
In this section, we demonstrate the merits of our sequential strategy through a variety of experiments. We also justify our design choices through extensive ablation studies. 
\subsection{Dataset}
Due to the unavailability of a large scale real-world dataset for this task,  we utilize the synthetic PartNet~\cite{mo2019partnet} dataset containing fine-grained instance segmentation. We use the three largest categories - i) chair, ii) table and iii) lamp with the predefined train (70\%), validation (10\%) and test (20\%) splits. Each shape contains 1000 points, sampled from part meshes using farthest point sampling. To ensure invariance to the rigid transformation of part point clouds, we transform them into their canonical space using PCA~\cite{pearson1901liii}.


\subsection{Loss Functions}
To explore structural variations, we incorporate the MoN loss~\cite{fan2017point}, along with random noise $\mathbf{z}_{j}$ in the initial hidden state. Considering our overall network as $f$ and the optimal pose-extractor as $f^{*}$,  we define the MoN loss in Equation \eqref{eq:mmd} as,
\begin{equation}~\label{eq:mmd}
\mathcal{L}_{mon} = \min _{j\in[N]} \mathcal{L}\left(f\left(\mathcal{P}, \mathbf{z}_{j}\right), f^{*}(\mathcal{P})\right).
\end{equation}
Here, $\mathbf{z}_{j} \sim \mathcal{N}(\mathbf{0},\mathbf{I}) \hspace{1mm} \forall \hspace{1mm}j \in[N]$, are IID random noise vectors. The loss function, $\mathcal{L}$, is split into three categories similar to~\cite{HuangZhan2020PartAssembly} for global and part-wise structural integrity.

Firstly, the translation is supervised by a Euclidean loss $\mathcal{L}_{t}$  (Equation \eqref{eq10}) between the predicted part center $\mathbf{c}_{i}$ and the ground-truth part center $\mathbf{c}_{i}^{*}$.
\begin{equation}
    \mathcal{L}_{t}=\sum_{i=1}^N\left\|\mathbf{c}_{i}-\mathbf{c}_{i}^{*}\right\|_{2}^{2}.
    \label{eq10}
\end{equation}
Secondly, the rotation is supervised by calculating Chamfer distance~\cite{fan2017point} between the rotated point cloud $\mathbf{q}_{i}(\mathbf{P}_{i})$ and the ground-truth point cloud $\mathbf{q}_{i}^{*}(\mathbf{P}_{i})$ (Equation \eqref{eq11}). 

\begin{eqnarray}
\mathcal{L}_r&=&\sum_{i=1}^Nd_c(\mathbf{q}_{i}(\mathbf{P}_{i}),\mathbf{q}_{i}^{*}(\mathbf{P}_{i})).
\label{eq11}
\end{eqnarray}
Here, $d_c(\mathcal{X},\mathcal{Y})$ is the Chamfer distance between the two point sets $\mathcal{X}$ and $\mathcal{Y}$, defined in Equation \eqref{eq14}.
\begin{equation}
d_c(\mathcal{X},\mathcal{Y})=\sum_{\mathbf{x} \in \mathcal{X}} \min _{\mathbf{y} \in \mathcal{Y}}\|\mathbf{x}-\mathbf{y}\|_{2}^{2}+ \sum_{\mathbf{y} \in \mathcal{Y}} \min _{\mathbf{x} \in \mathcal{X}}\|\mathbf{x}-\mathbf{y}\|_{2}^{2}.
\label{eq14}
\end{equation}
Lastly, the shape-cd-loss $\mathcal{L}_{s}$ (Equation \eqref{eq:scd}), ensures the overall quality of the generated assembly $\mathcal{S}$ by computing its Chamfer distance from the ground truth assembly $\mathcal{S}^{*}$. 
\begin{equation}~\label{eq:scd}
    \mathcal{L}_{s}=d_c(\mathcal{S},\mathcal{S}^*).
\end{equation}


\begin{figure}[!h]
    \centering
    \includegraphics[width = 1\linewidth]{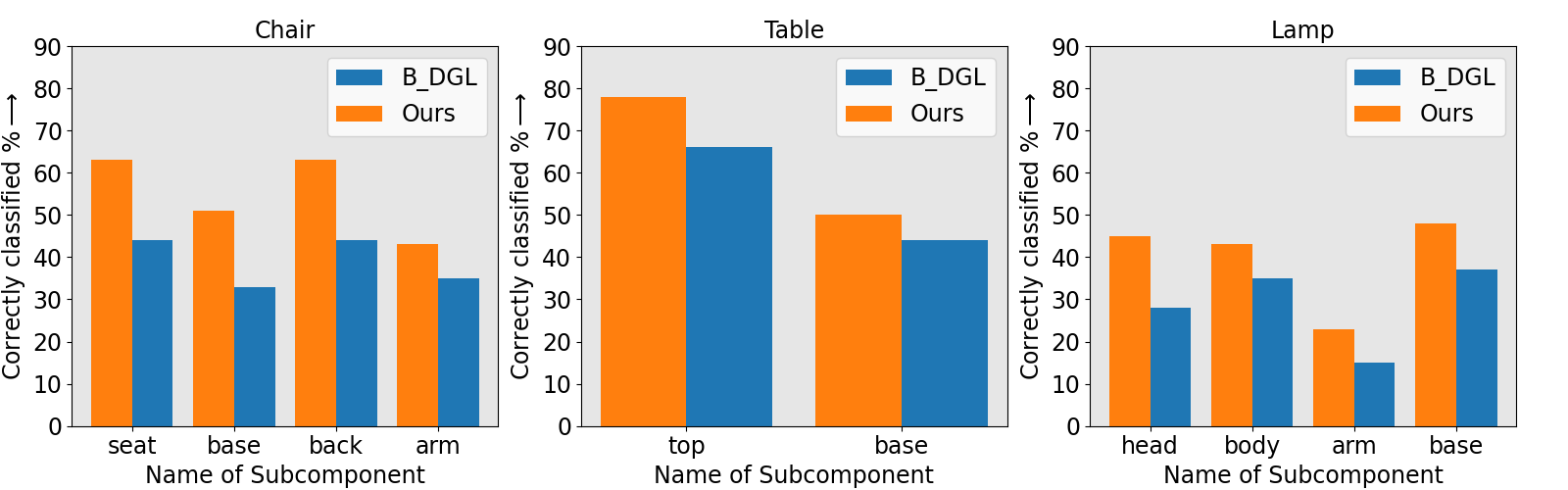}
    \caption{Comparison of our method with B-DGL~\cite{HuangZhan2020PartAssembly} on the most common sub-components of each category.}
    \label{fig:subcomponent_acc}
\end{figure}

\setlength{\tabcolsep}{1pt}	
	\begin{figure*}[h]
	\begin{minipage}{0.7\textwidth} 
		\centering
		\begin{tabular}{cccc|ccc|ccc}
		  \raisebox{3.5\height}{B-Global~\cite{schor2019componet, li2020pagenet}}	
			 &
			\includegraphics[width=0.13\linewidth]{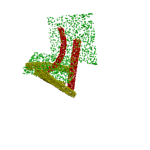}
			&\includegraphics[width=0.13\linewidth]{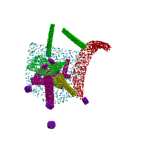}
			&\includegraphics[width=0.13\linewidth]{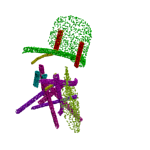}
			&\includegraphics[width=0.13\linewidth]{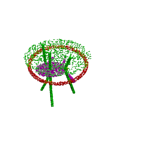}
			&\includegraphics[width=0.13\linewidth]{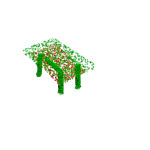}
			&\includegraphics[width=0.13\linewidth]{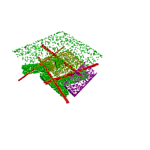}
			&\includegraphics[width=0.13\linewidth]{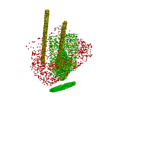}
			&\includegraphics[width=0.13\linewidth]{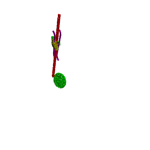}
			&\includegraphics[width=0.13\linewidth]{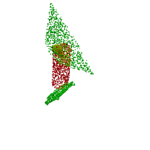}
			\\
			\raisebox{3.5\height}{B-LSTM~\cite{wu2020pq}} & \includegraphics[width=0.13\linewidth]{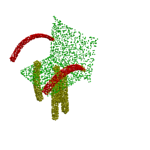}
			&\includegraphics[width=0.13\linewidth]{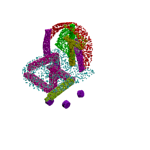}
			&\includegraphics[width=0.13\linewidth]{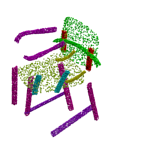}
			&\includegraphics[width=0.13\linewidth]{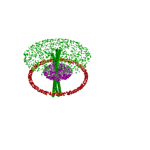}
			&\includegraphics[width=0.13\linewidth]{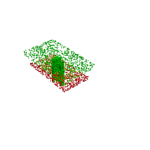}
			&\includegraphics[width=0.13\linewidth]{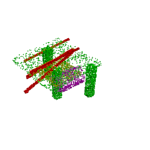}
			&\includegraphics[width=0.13\linewidth]{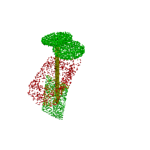}
			&\includegraphics[width=0.13\linewidth]{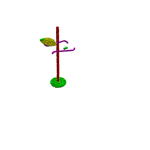}
			&\includegraphics[width=0.13\linewidth]{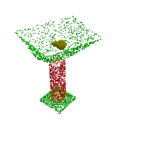}
			\\
			\raisebox{3.5\height}{B-DGL~\cite{HuangZhan2020PartAssembly}} & \includegraphics[width=0.13\linewidth]{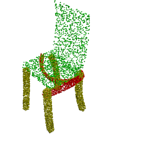}
			&\includegraphics[width=0.13\linewidth]{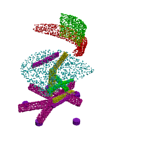}
			&\includegraphics[width=0.13\linewidth]{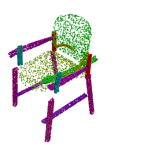}
			&\includegraphics[width=0.13\linewidth]{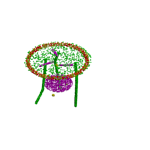}
			&\includegraphics[width=0.13\linewidth]{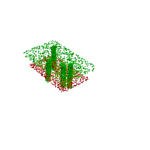}
			&\includegraphics[width=0.13\linewidth]{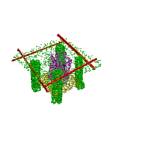}
			&\includegraphics[width=0.13\linewidth]{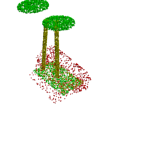}
			&\includegraphics[width=0.13\linewidth]{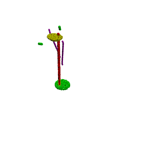}
			&\includegraphics[width=0.13\linewidth]{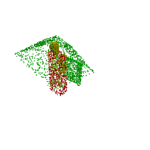}
			\\
			\raisebox{3.5\height}{Ours} & 
			\includegraphics[width=0.13\linewidth]{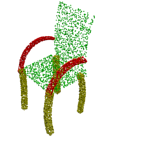}
			&\includegraphics[width=0.13\linewidth]{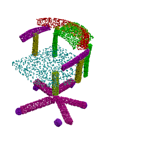}
			&\includegraphics[width=0.13\linewidth]{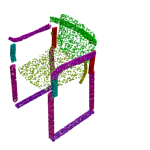}
			&\includegraphics[width=0.13\linewidth]{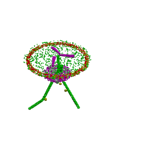}
			&\includegraphics[width=0.13\linewidth]{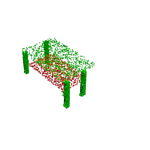}
			&\includegraphics[width=0.13\linewidth]{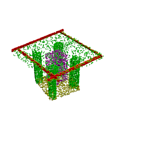}
			&\includegraphics[width=0.13\linewidth]{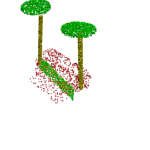}
			&\includegraphics[width=0.13\linewidth]{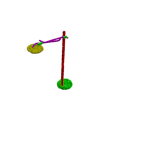}
			&\includegraphics[width=0.13\linewidth]{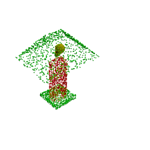}
			\\
			\raisebox{3.5\height}{Ground Truth} & 
			\includegraphics[width=0.13\linewidth]{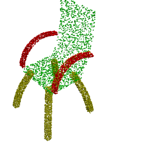}
			&\includegraphics[width=0.13\linewidth]{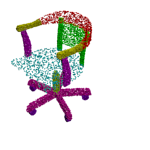}
			&\includegraphics[width=0.13\linewidth]{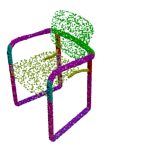}
			&\includegraphics[width=0.13\linewidth]{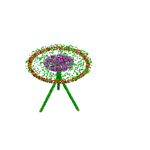}
			&\includegraphics[width=0.13\linewidth]{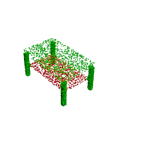}
			&\includegraphics[width=0.13\linewidth]{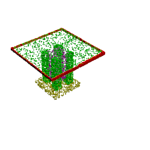}
			&\includegraphics[width=0.13\linewidth]{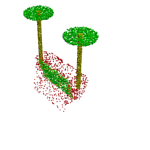}
			&\includegraphics[width=0.13\linewidth]{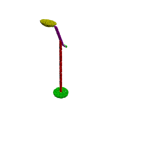}
			&\includegraphics[width=0.13\linewidth]{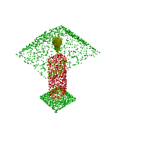}
			\\
			& (a) & (b) & (c) & (d) & (e) & (f) & (g) & (h) & (i)
			\\
			& \multicolumn{3}{c}{Chair} &
			\multicolumn{3}{c}{Table} &
			\multicolumn{3}{c}{Lamp} 
		\end{tabular}
		\end{minipage}%
		\caption{Qualitative comparison with baseline methods on 9 different shapes, (a)-(i) of PartNet~\cite{mo2019partnet}.}
		\label{fig:complete}
	\end{figure*}
	

\subsection{Evaluation Metrics}
We measure the network performance by generating a variety of shapes and finding the closest shape to the ground truth using minimum matching distance~\cite{achlioptas2018learning}. For better comparison, we utilize \textit{part accuracy}, \textit{connectivity accuracy} and \textit{shape Chamfer distance}, used by ~\cite{HuangZhan2020PartAssembly}. 
\textit{Shape Chamfer distance} is defined in Equation~\eqref{eq:scd}. We define the remaining terms below. 

\textbf{Part Accuracy.} This metric (Equation~\eqref{eq16}) measures the fraction of $SE(3)$ transformed parts $\mathbf{T}_{i}(\mathbf{P}_{i})$ that lie below a threshold Chamfer distance $\tau_{p}$ from the ground truth $\mathbf{T}_{i}^{*}(\mathbf{P}_{i})$. Here, $\mathbb{1}$ represents the indicator function.
\begin{equation}
PA = \frac{1}{N} \sum_{i=1}^N\mathbb{1}\big(d_c(\mathbf{T}_{i}\left(\mathbf{P}_{i}\right),\mathbf{T}_{i}^{*}\left(\mathbf{P}_{i}\right))<\tau_p\big).
\label{eq16}
\end{equation}

\textbf{Connectivity Accuracy.} We incorporate connectivity accuracy (Equation \eqref{eq17}), to measure the quality of inter-part connections. For each connected-part pair ($\mathbf{P}_{i}, \mathbf{P}_{j}$), we define the contact $c_{ij}^{*}$ as a point on $\mathbf{P}_{i}$ that is closest to $\mathbf{P}_{j}$. Similarly, contact point $c_{ji}^{*}$ is the point on $\mathbf{P}_{j}$ that is closest to $\mathbf{P}_{i}$. $(c_{ij}^{*}, c_{ji}^{*})$ are transformed into their corresponding part canonical space as $(c_{ij}, c_{ji})$. Then, connectivity accuracy is calculated as,  
\begin{equation}
CA = \frac{1}{|\mathcal{C}|} \sum_{\left\{c_{i j}, c_{j i}\right\} \in \mathcal{C}} \mathbb{1}\left(\left\|\mathbf{T}_{i}\left(c_{i j}\right)-\mathbf{T}_{j}\left(c_{j i}\right)\right\|_{2}^{2}<\tau_{c}\right).
\label{eq17}
\end{equation}
Here, $\mathcal{C}$ represents the set of all possible contact point pairs $\{c_{ij}, c_{ji}\}$. During evaluation, $\tau_{c}$ and $\tau_{p}$ are set to 0.01.
\subsection{Results and Comparisons}
The only direct baseline to our work is Dynamic Graph Learning              (B-DGL)~\cite{HuangZhan2020PartAssembly}. We also compare our results with three other baselines: B-LSTM~\cite{wu2020pq}, B-Global~\cite{li2020pagenet, schor2019componet} and B-Complement~\cite{sung2017complementme} used by B-DGL. As we were unable to reproduce the results of B-Complement accurately, we exclude it from our qualitative comparison.

In Table~\ref{table:performance}, we observe that the most improvement in part accuracy($\approx10\%$) occurs in the chair category. This could be due to four distinct components of the chair - back, seat, leg and arm, which merits our progressive assembly framework. The improvement is $\approx4.5\%$ on the table category, which has only two such distinct components - table-top and table-base.  
On the lamp category, progressive assembly helps to ensure better connectivity accuracy, which is $15\%$ above B-DGL.  
\par
Figure~\ref{fig:subcomponent_acc} shows that the improvement is distributed across the most common subcategories of a shape. Among these, the chair-seat, chair-back and table-top are well-performing subcategories. On the other hand, structurally diverse components like the chair arm and table base have lower accuracy's.

Our qualitative results reflect a few key aspects which our progressive framework improves. We further highlight these qualitative results in our supplementary video.  

\textbf{Inter-Part Confusion.}  In Figure~\ref{fig:complete}(a), we observe that the chair arm and leg share a very similar structure. B-DGL misinterprets the curved red chair handles. Our framework is able to correctly place this part. 

\textbf{Better Connectivity.} Our method better understands fine-grained part connections. In Figure~\ref{fig:complete}(f), our network is able to correctly predict the four bars around the table-top. In the lamp in Figure~\ref{fig:complete}(i), our network is able to  predict the light bulb and its cover correctly. 
\par
\textbf{Rotational Symmetry.} Predicting rotational symmetry is a challenging task which our network handles very well. In Figure~\ref{fig:complete}(b) the star shaped chair legs are correctly predicted. 

\renewcommand{\arraystretch}{1}
\begin{figure}
    \centering
    \includegraphics[width = 1\linewidth]{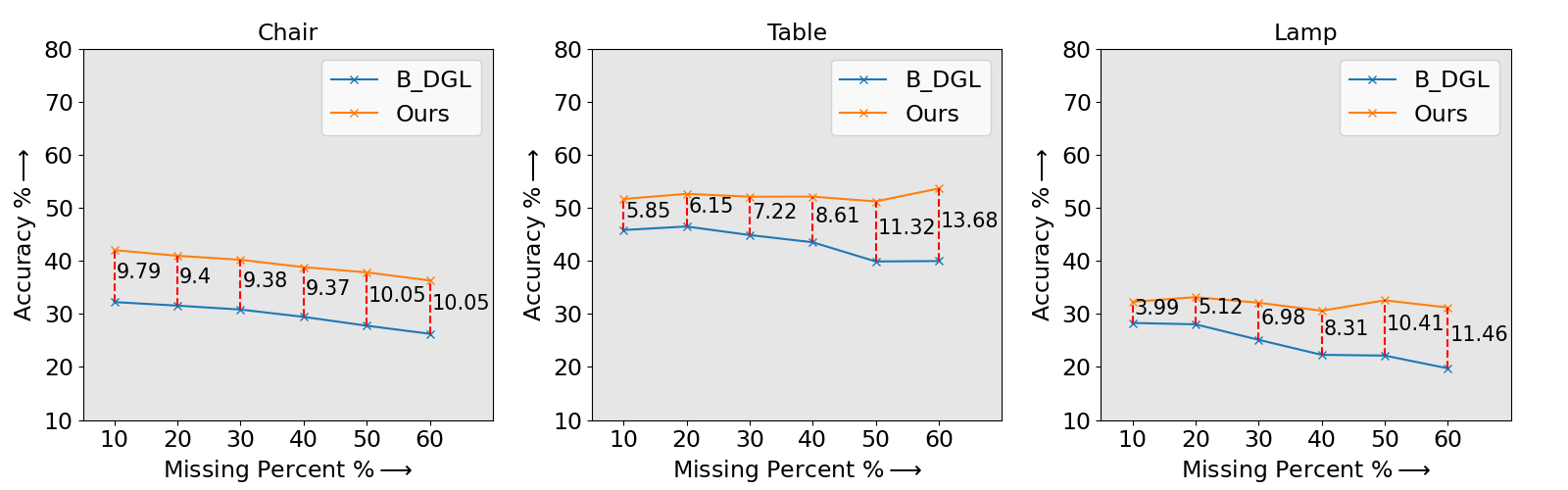}
    \caption{Comparison of our method with B-DGL~\cite{HuangZhan2020PartAssembly} with varying percentage of missing parts. }
    \label{fig:missing_acc}
\end{figure}
\subsection{Performance with Missing Parts}
Often, a packaging defect can result in missing parts during assembly. In this scenario, we want our algorithm to predict plausible results so the deficiency can be identified. 
\par
This is not without a few caveats. By randomly choosing a candidate for deletion,  pivotal parts like the chair-seat could be removed, affecting the quality of assembly. Instead, we order parts according to their volume and delete a certain percentage of the smallest volume parts. We utilize this strategy as smaller parts are more likely to be misplaced.
\begin{figure}[!h]
    \begin{center}
    \stackunder{\includegraphics[width=0.25\linewidth]{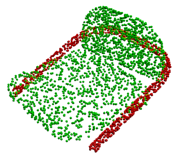}}{B-DGL~\cite{HuangZhan2020PartAssembly}}
    \hspace{4mm}
    \stackunder{\includegraphics[width=0.25\linewidth]{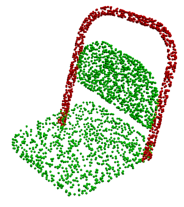}}{Ours}
    \hspace{4mm}
    \stackunder{\includegraphics[width=0.25\linewidth]{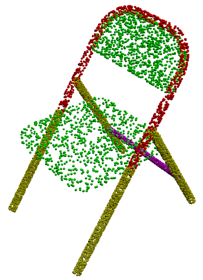}}{Ground Truth}
    \end{center}    
\caption{Sample result with missing chair legs. Notice how our method approximately positions the curved back.}
\label{fig:shape-missing}
\end{figure}

We compute the volume of a point cloud as the volume of its axis-aligned-bounding-box. Further, each part belonging to a part-group (ex:-chair legs) is assigned the minimum volume among all its members. This accounts for point-cloud sampling variations. Moreover, we do not use partial part-groups. For instance, if we delete one chair leg, the rest of the chair legs are also deleted. 
\par
In Figure~\ref{fig:missing_acc}, we observe that the accuracy increases at specific deletion percentages. This could be due to the removal of incorrectly placed smaller volume parts. The increasing difference with the baseline B-DGL shows that our algorithm is more robust at higher deletion percentages. In Figure~\ref{fig:shape-missing}, we provide a sample visual result on a chair at 60\% deletion.

\subsection{Shape Recovery from Latent Space}

An exciting application of our latent space is shape recovery. Unlike the task of shape-autoencoding~\cite{yang2018foldingnet, chen2019bae, pang2020tearingnet}, we do not have access to the global shape structure. Instead, we recover shape structure from the component point clouds. We utilize the point cloud decoder of TreeGAN~\cite{shu20193d} and train it without the discriminator separately on the two largest categories, chair and table, using the last hidden state of our GRU.  We optimize reconstruction loss using \textit{shape Chamfer distance} (Equation~\eqref{eq:scd}) and train it independently of our assembly framework. More details of our training strategy are included in the supplementary file.
\par
In Figure~\ref{fig:shape-recovery}, we observe that for the four-leg table, our recovery maintains structural integrity. On the chair, our network gives a good outline of the structure, with a sparser distribution of points around the chair legs. This loss of information could be due to storing a detailed shape in a smaller dimension.
\par
\begin{figure}[!h]
    \begin{center}
    \stackunder{\includegraphics[width=0.17\linewidth]{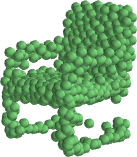}}{Prediction}
    \stackunder{\includegraphics[width=0.185\linewidth]{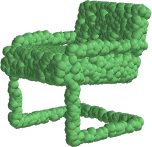}}{Ground-truth}
    \stackunder{\includegraphics[width=0.27\linewidth]{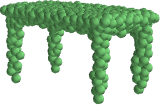}}{Prediction}
    \stackunder{\includegraphics[width=0.27\linewidth]{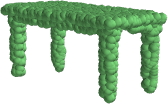}}{Ground-truth}
    \end{center}    
\vspace{-0.4cm}
\caption{Results on shape recovery from hidden state. Our method can recover coarse structure of the shape. }
\label{fig:shape-recovery}
\end{figure}

This experiment gives an insight into our progressive scheme. Our latent state carries coarse structure information required in the subsequent steps. The reconstruction is reasonably accurate considering that the hidden state has not been constrained during assembly and the structural diversity of PartNet~\cite{mo2019partnet}. 


\begin{figure}[t!]
    \centering
    {
    \includegraphics[width = 1\linewidth]{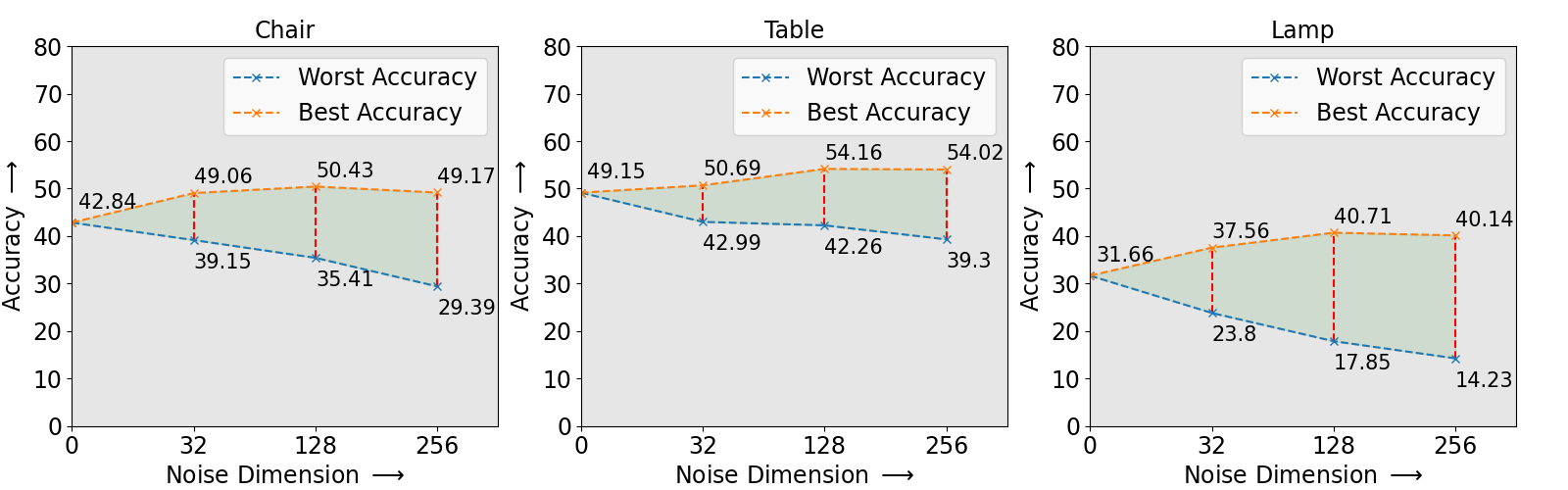}
    }
    \caption{Performance of our network on varying dimension of random noise. Our results at zero noise are comparable to B-DGL~\cite{HuangZhan2020PartAssembly} (Chair: $39.00$, Table: $49.51$, Lamp: $33.33$).}
    \label{fig:variation}
\end{figure}

\subsection{Bounds of Performance}
Introducing random noise to our network architecture allows us to generate structural variations. In this section, we monitor the performance of our network to varying amounts of randomness and establish a bound on part accuracy.  We do not modify the network architecture but change the dimension of random noise in Equation~\eqref{eq:initial_hidden}, keeping the dimension of the hidden state $\mathbf{h}_{1}^{(t)}$ and $\mathbf{g}_{N}^{(t)}$ fixed. 

To better quantify the variations, we introduce the term variability $V_{E}$ as the difference between its best and worst performance. Analogous to Equation~\eqref{eq:mmd}, we define maximum matching distance as the worst performance of our network over $E$ iterations. Then, considering our network as a function, $f$, the ground truth as $f^{*}$, and random noise as a vector $\mathbf{z}_{j}$,
\begin{equation}
\begin{array}{r}
V_{E}=\max _{j \in[E]} \mathcal{L}\left(f\left(\mathcal{P}, \mathbf{z}_{j}\right), f^{*}(\mathcal{P})\right)- \\
\min _{j \in[E]} \mathcal{L}\left(f\left(\mathcal{P}, \mathbf{z}_{j}\right), f^{*}(\mathcal{P})\right).
\end{array}
\end{equation}
To replicate a practical scenario, we choose $E= 10$ for this analysis. We experiment with noise dimensions of $0$, $32$, $128$ and $256$ and report part-accuracy on each category. 
\par
Our results in Figure~\ref{fig:variation} demonstrate that increasing the random noise allows the network to explore more structural varieties; however, it results in a decreasing infimum. Also, at a given noise dimension, the lamp category shows the highest variability. This could be attributed to its structural diversity and smaller dataset size. 
\par
We customize our network design choices based on this analysis.  For optimal performance, our network must balance accuracy ($PA\uparrow$) and variability ($V_{E}\downarrow$). Accordingly, we choose the noise dimension as $32$ for the chair and lamp category and $128$ for the table category.

\textbf{Performance in Absence of Random Noise.}
Incorporating MoN~\cite{fan2017point} loss during training allows exploration of structural varieties, leading to better overall performance. However, it comes at the cost of increased training time and variability ($V_{E}$). Figure~\ref{fig:variation} reflects an additional benefit of our progressive scheme; our results are competitive even if no random noise is incorporated. In this setting, we can train our network without including MoN loss, which is $\times2.5$ faster and has no tradeoff on variability ($V_{E} = 0$). In Table~\ref{table:performance}, we observe that these results are comparable to B-DGL trained with $5$ iterations of MoN~\cite{fan2017point}.

\subsection{Ablation Studies}
In this section, we provide an experimental justification of our design choices. In particular, we consider two major aspects - i) structural variations of our architecture and ii) optimal sequence for assembly. We provide details of each configuration in our supplementary file. 

\textbf{Architectural Variants}. We construct a diverse set of architecture variants to justify our design choices. We use a unidirectional RNN in both i) bottom to top and ii) top to bottom ordering, iii) we initialize the subsequent hidden state, ($\mathbf{h}_{1}^{(t+1)} = \mathbf{h}_{N}^{(t)}$ and $\mathbf{g}_{N}^{(t + 1)} =  \mathbf{g}_{1}^{(t)}$), iv) we add noise to the pose decoder instead of the hidden state, v) we evaluate our recurrent backbone without graph learning, and vi) we pass the precomputed graph-message after sequential encoding.  

In Table~\ref{table:struct-variations}, we observe that the bidirectional GRU incorporates more context compared to its unidirectional counterpart. Interestingly, using bottom-to-top encoding performs better ($PA = 46.42$) than top-to-bottom ($PA = 44.81$) encoding. One reason for this could be that the chair legs are closer to the seat, and fixing the seat location earlier in the sequence can better predict the chair arm and back. 

Our standalone framework can predict parts better ($PA = 45.36$) than the B-DGL  ($PA = 39.00$), highlighting the merits of progressive assembly.  It is  noteworthy to observe that initializing hidden states of the subsequent time-steps $t > 1$ negatively impacts part accuracy ($PA = 46.74$). 
This could be because using random noise at each step better explores structural variations than initializing them with the previous hidden state. Also, exploring global structural variations by introducing the noise in the hidden state ($PA = 49.06$) results in better performance than part-wise randomness, i.e, placing noise in the pose-decoder ($PA = 46.31$).

We also analyze the importance of different loss functions by removing each separately and training with the remaining losses. In Table~\ref{tab:loss-variants}, we observe that $\mathcal{L}_{t}$ is the most significant for accurate part placement. Among the remaining losses, $\mathcal{L}_{r}$ helps improve connectivity between parts ($CA$), and $\mathcal{L}_{s}$ helps optimize the overall shape structure ($SCD$).
\setlength{\tabcolsep}{2pt}
\begin{table}[t]\label{tab:ablation}
	\small
	\begin{center}
		\begin{tabular}{c|c|c|c}
			\hline
			& \hspace{3mm} SCD $\downarrow$\hspace{3mm} & \hspace{3mm} PA $\uparrow$\hspace{3mm} & \hspace{3mm} CA $\uparrow$\\
			\hline
		    (i) Bottom to Top Encoding & \textbf{0.0086} & 46.42 & 29.66 \\
			\hline				
			(ii) Top to Bottom Encoding & 0.0101 & 44.81 & 28.85\\
			\hline
			(iii) Initialize hidden states & 0.0095 & 46.74 & 29.60\\
			\hline
		    (iv) Noise in Pose Decoder & 0.0098 & 46.31 & 31.19 \\
			\hline
			(v) Without Graph Learning & 0.0092 & 45.36 & 31.78\\
			\hline
			(vi) Sequential before Graph & 0.0091 & 48.13 & 
		    30.54 \\
			\hline
		    (vii) Ours (Complete) & 0.0087 &\textbf{49.06} & \textbf{32.26}\\
			\hline
		\end{tabular}
	\end{center}
	\caption{Ablation study of structural variants. Here, SCD: Shape Chamfer Distance.}
	\label{table:struct-variations}
\end{table}
\setlength{\tabcolsep}{2pt}
\begin{table}[!h]
	\small
	\begin{center}
		\begin{tabular}{c|c|c|c}
			\hline
			&\hspace{4mm} SCD $\downarrow$ \hspace{3mm} &\hspace{3mm} PA $\uparrow$ \hspace{3mm}& \hspace{2.5mm} CA $\uparrow$\\
			\hline				
			(i) Without $\mathcal{L}_{s}$ & 0.0098 & 48.62 & 30.85\\
			\hline
			(ii) Without $\mathcal{L}_{t}$ & 0.0091 & 16.35 & 14.21 \\
			\hline
			(iii) Without $\mathcal{L}_{r}$ & \textbf{0.0078} & 48.72 & 29.85\\
			\hline
			(iv) Ours (Complete)  & 0.0087 & \textbf{49.06} & \textbf{32.26}\\
			\hline
		\end{tabular}
	\end{center}
	\caption{Removing individual loss functions. Here, SCD: Shape Chamfer Distance. }
	\label{tab:loss-variants}
\end{table}

\textbf{Optimal Order for Assembly.}
As our assembly strategy is progressive, studying the interplay between ordering and the resulting part placement is crucial. However, the number of possible arrangements grows exponentially with the number of parts. Theoretically, there could exist an order which produces better assembly results than ours. Identifying this global optimum ordering is beyond the scope of this experiment. Instead, we consider a few other intuitive choices and determine the best one among those - i) we consider volume ordering, i.e., parts ordered from minimum to maximum volume,
ii) we group similar parts together, start from a random group and iteratively append neighbouring groups, iii) we start from a random part and iteratively append neighbouring parts (part-connectivity), iv) we follow part-connectivity, however, beginning at the part with maximum neighbours, v) and lastly, we evaluate random ordering. 
\par 
The results in Table~\ref{tab:order-ablation} show that among our considered choices, the top-down ordering ($CA = 32.26$) of parts is optimal for training, and random arrangement performs the worst ($CA = 19.04$).  
Among the other choices, part connectivity ensures better connectivity ($CA = 25.19$) compared to group-wise ordering ($CA = 22.07$). Moreover, starting from the most connected part further improves connectivity accuracy ($CA=28.65$). However, there is not much difference in following volume ordering ($CA = 22.01$) and group connectivity ordering ($CA = 22.07$). 

\setlength{\tabcolsep}{2pt}
\begin{table}[t]
	\small
	\begin{center}
		\begin{tabular}{c|c|c|c}
			\hline
			&\hspace{4mm} SCD $\downarrow$ \hspace{3mm} &\hspace{3mm} PA $\uparrow$ \hspace{3mm}& \hspace{2.5mm} CA $\uparrow$\\
			\hline				
			(i) Volume order & 0.0119 & 36.13 & 22.01\\
			\hline
			(ii) Group Connectivity Order & 0.0118 & 36.62 & 22.07 \\
			\hline
			(iii) Part Connectivity Order & 0.0114 & 37.46 & 25.19 \\
			\hline
			(iv) Central - Part Connectivity & 0.0102 & 43.04 & 28.65 \\
			\hline
		    (v) Random order & 0.0158 & 30.91 & 19.04\\
		    \hline
			(vi) Top to Bottom Order &\textbf{0.0087} &\textbf{49.06} & \textbf{32.26}\\
			\hline
		\end{tabular}
	\end{center}
	\caption{Ablation study of the different orders used for assembling a shape.  Here, SCD: Shape Chamfer Distance. }
	\label{tab:order-ablation}
\end{table}

\section{Conclusion and Future Work}
We proposed a novel progressive approach to assemble shapes given their part point clouds which can better predict part locations and inter-part connectivity.  We showed the potency of our latent space by utilizing it to recover shape structure. Furthermore, our experiments demonstrated that part ordering could play a crucial role in assembly automation. Future works may develop a reordering framework to arrange randomly-ordered parts into a consistent top to bottom order. Another possible study could explore the application of our latent space in retrieving part-connectivity. 

We would also like to incorporate constraints such as symmetry into our progressive strategy. In a chair, we could assemble its back, followed by its seat and legs ensuring global shape symmetry constraints at each step. This would reduce the dimensionality of the assembly space.    

{\small
\bibliographystyle{ieee_fullname}
\bibliography{paper_update}
}
\end{document}